\documentclass[10pt,twocolumn,letterpaper]{article}

\usepackage{cvpr}
\usepackage{cuted} %< for strip environment

\usepackage{currfile} %< for \currfilebase macro

\usepackage{caption} %< for "\captionof" commands (teaser)
\usepackage{booktabs}
\usepackage{multirow}
\usepackage{graphicx}
\usepackage{amsmath,amssymb}
\usepackage{placeins}

\definecolor{cvprblue}{rgb}{0.21,0.49,0.74}
\usepackage[pagebackref,breaklinks,colorlinks,allcolors=cvprblue]{hyperref}
\usepackage{orcidlink}

\def\paperID{*****}
\def\confName{CVPR}
\def\confYear{2027}

\newcommand{\gl}{\texttt{global\_last}}

\title{Does VGGT Know What It Does Not Know?}

\author{Adityan Thirumani Srinivasan\,\orcidlink{0009-0002-9911-0586}\\
University of Wisconsin--Madison\\
\and
David Nordstr{\"o}m\,\orcidlink{0009-0001-4460-1508}\\
Chalmers University of Technology\\
}

\begin{document}
\twocolumn[{%
\renewcommand\twocolumn[1][]{#1}%
\maketitle
\vspace{-1.2em}
\begin{center}
  \centering
  \includegraphics[width=\textwidth]{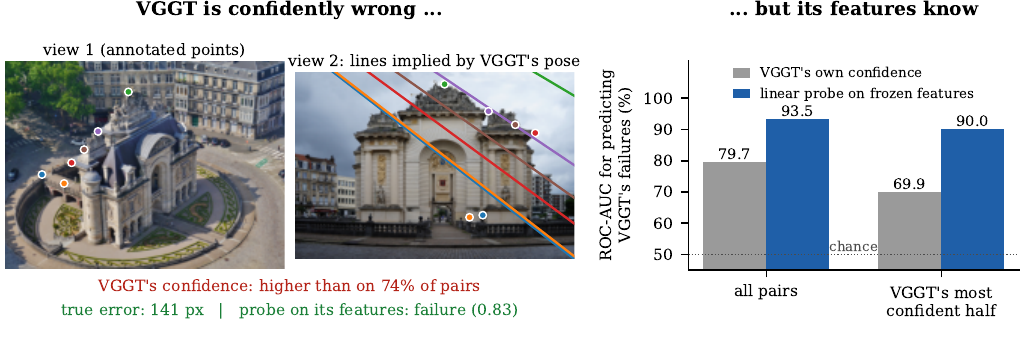}
  \captionof{figure}{\textbf{VGGT is confidently wrong, but its features know.} Left: an aerial and a ground view of the same gate. VGGT is more confident on this pair than on $74\%$ of HardMatch pairs, yet the epipolar lines implied by its predicted cameras miss the annotated points by $141$\,px; a linear probe on its frozen features flags the failure. Right: ROC-AUC for predicting VGGT's pose failures on held-out HardMatch categories, from VGGT's own confidence and from the probe, on all pairs and on the half of the pairs VGGT is most confident about.}
  \label{fig:teaser}
\end{center}
\vspace{0.6em}
}]

\begin{abstract}
Feed-forward 3D reconstruction models such as VGGT regress cameras and geometry in a single forward pass, but they report no uncertainty for the camera poses they predict: an incorrect relative pose looks no different from a correct one.
We ask whether the frozen model nonetheless \emph{knows} when its pose is wrong.
On HardMatch, a benchmark of 1{,}000 deliberately difficult image pairs on which VGGT fails on 35\% of pairs, a linear probe on VGGT's frozen features predicts pose failure on held-out categories with ROC-AUC 93.5\%, compared with 79.7\% for VGGT's own point-map confidence and 77.7\% for a classifier trained on all of VGGT's outputs.
The probe catches failures VGGT is confident about, lets a downstream system abstain on the pairs it distrusts, and transfers without retraining to rotation failures on PhotoTourism and ScanNet-1500, beyond what co-visibility explains.
This knowledge is absent from the per-image encoder: it is computed partway through the layers that fuse the two views, and the same holds for a second model, $\pi^3$.
In essence, we find that VGGT knows more about its own failures than it reports.
\end{abstract}

\section{Introduction}
\label{sec:intro}

Feed-forward reconstruction models~\cite{wang2024dust3r,leroy2024grounding,wang2025vggt,wang2026vggtomega,wang2026pi,depthanything3,keetha2025mapanything} predict cameras, depth and point maps from a set of images in a single forward pass, replacing much of the classical Structure-from-Motion (SfM)~\cite{hartley2003multiple} pipeline.
VGGT~\cite{wang2025vggt} is the most widely used of these models.
It outputs a per-pixel confidence for depth and point maps, but no uncertainty for the camera poses it regresses. Downstream systems, such as SLAM, mapping, novel-view synthesis, therefore cannot distinguish a reliable relative pose from a badly wrong one.

Two observations make this gap worth studying.
First, these models do fail: on HardMatch~\cite{nordstrom2026loma}, a recent benchmark of hand-annotated, deliberately difficult image pairs, VGGT's predicted geometry is grossly wrong (median epipolar error above 50\,px) on 35\% of pairs.
Second, the large-language-model literature has shown that internal representations often encode whether a model's own output is correct~\cite{kadavath2022language,burns2023discovering,azaria2023internal}, even when the model's stated confidence does not.
This motivates the question in our title: \emph{does a feed-forward 3D model know when its pose is wrong, and is that knowledge linearly readable from its frozen features?}

We answer it with linear probes~\cite{alain2017,belinkov2022probing} on frozen VGGT activations at every block, evaluated with category-grouped cross-validation so that every test category is unseen during training, and compared against everything VGGT itself outputs. \noindent\textbf{The main contributions are as follows:}
\begin{itemize}\itemsep1pt
  \item We show that VGGT's frozen features encode its own pose failures far more clearly than anything it outputs (ROC-AUC $93.5\%$ \vs $79.7\%$ for its own confidence on HardMatch; \cref{fig:teaser}), including failures it is confident about.
  \item We locate where this knowledge forms: it is absent from the per-image encoder and emerges partway through the layers that fuse the views, in VGGT and in a second model, $\pi^3$.
  \item We show that the signal is useful beyond the benchmark it was trained on: it supports selective prediction and transfers without retraining to rotation failures on PhotoTourism and ScanNet-1500.
\end{itemize}

\section{Related Work}
\label{sec:related}

\paragraph{Feed-forward 3D reconstruction.}
DUSt3R~\cite{wang2024dust3r} regresses aligned point maps for an image pair and MASt3R~\cite{leroy2024grounding} adds dense matching. VGGT~\cite{wang2025vggt} predicts cameras, depth, point maps and tracks for many views in one pass through alternating frame-wise and global attention; VGGT-$\Omega$~\cite{wang2026vggtomega}, $\pi^3$~\cite{wang2026pi}, Depth Anything 3~\cite{depthanything3} and MapAnything~\cite{keetha2025mapanything} extend the recipe. These models expose per-pixel confidence for geometry, learned with an uncertainty-weighted loss, but no uncertainty for camera poses.

\paragraph{Uncertainty of feed-forward reconstruction.}
Hillemann \etal~\cite{hillemann2026uncertainty} evaluate how well VGGT's per-pixel confidence tracks depth and point errors on DTU and note that VGGT provides no pose uncertainty. ReVar3R~\cite{mahdi2026revar3r} estimates per-point uncertainty by perturbing the inputs of a frozen model. Both target per-point geometry; we target the correctness of the predicted \emph{relative pose}, and we read it from internal features rather than outputs or input perturbations.

\paragraph{What 3D foundation models encode.}
Zhang \etal~\cite{zhang2026emergentextremeviewgeometry3d} find an emergent ability to reason about non-overlapping views; Star{\'y} \etal~\cite{stary2025understanding} trace how relative pose forms across the layers of a DUSt3R variant; Nordstr{\"o}m \etal~\cite{nordstrom2026romaomega} study what feed-forward 3D models know about image matching. Closest to us, Co-VGGT~\cite{ziliotto2026covggt} probes VGGT's layers for \emph{co-visibility}. Co-visibility is a property of the scene; failure is a property of the prediction. HardMatch pairs are co-visible by construction, and on PhotoTourism our probe predicts failure beyond ground-truth co-visibility (\cref{sec:pt}).

\paragraph{Probing, self-knowledge and failure prediction.}
Linear probes measure what is linearly available in frozen features~\cite{alain2017,belinkov2022probing}. For language models, internal states encode whether the model's own answer is correct~\cite{kadavath2022language,burns2023discovering,azaria2023internal}. Failure prediction and selective prediction~\cite{hendrycks2017baseline,geifman2017selective,kendall2017uncertainties} typically train dedicated heads; we keep the model frozen and compare against the confidence it already emits.

\section{Method}
\label{sec:method}

\paragraph{Model and inputs.}
We use VGGT~\cite{wang2025vggt} (1B) frozen, with no fine-tuning. Each image pair is processed in one forward pass of two frames; we study pairs because HardMatch defines its ground truth, and its official evaluation, per pair, and test up to $16$ views in \cref{sec:pt}. Images are passed as RGB in $[0,1]$ (VGGT normalises internally), resized so that the longer side is $518$\,px with the shorter side a multiple of $14$, and centred on a square canvas. From the predicted pose encoding we decode extrinsics $[\mathbf{R}_i|\mathbf{t}_i]$ and intrinsics $\mathbf{K}_i$ for both frames. As a second model we use $\pi^3$~\cite{wang2026pi} with the same inputs; it predicts camera-to-world poses but no intrinsics, so we fit a focal length by least squares to its local point maps. We hook its encoder blocks $6$, $12$ and $23$, all $36$ decoder blocks and the camera head's features.

\subsection{Defining failure}
\paragraph{Pairs without ground-truth poses (HardMatch).}
HardMatch~\cite{nordstrom2026loma} provides, for each pair, a set of hand-annotated correspondences (median $11$ per pair) but no camera poses. Following the official protocol, which evaluates any method through the fundamental matrix it implies, we form VGGT's own relative pose $\mathbf{R}=\mathbf{R}_2\mathbf{R}_1^{\top}$, $\mathbf{t}=\mathbf{t}_2-\mathbf{R}\mathbf{t}_1$ and fundamental matrix
\begin{equation}
  \mathbf{F} = \mathbf{H}_2^{\top}\,\mathbf{K}_2^{-\top}[\mathbf{t}]_{\times}\mathbf{R}\,\mathbf{K}_1^{-1}\,\mathbf{H}_1 ,
\end{equation}
where $\mathbf{H}_i$ maps original-image pixels to the network's input canvas. For every annotated correspondence we compute the symmetric epipolar distance~\cite{mishkin2015WXBS}, take the median over the pair, and call the pair a \emph{failure} if it exceeds $\tau$. We use $\tau=50$\,px (about $5\%$ of the image width) as the main threshold and report $10$, $20$ and $100$\,px. As in the official code, we keep pairs with at least $8$ correspondences and drop pairs labelled \emph{dynamic}, leaving $903$ pairs in $97$ categories.

\paragraph{Pairs with ground-truth poses (PhotoTourism, ScanNet-1500).}
Here we compare VGGT's relative pose with ground truth directly: the geodesic rotation error, the angle between translation directions (sign-agnostic, as in the ScanNet-1500 protocol~\cite{sun2021loftr}), and the pose error $\max(e_R, e_t)$.

\subsection{Probing for failure}
\paragraph{Features.}
We hook seven depths of the frozen network: blocks $6$, $12$ and $23$ of the DINOv2-initialised~\cite{oquab2023dinov2} ViT encoder, which processes each image independently; the output of the aggregator's last frame-attention block and last global (cross-frame) attention block; the normalised camera token that enters the camera head; and the input of the camera head's trunk at its final refinement iteration. For the layer analysis we additionally pool the output of every one of the $24$ encoder and $48$ aggregator blocks. Tokens are mean-pooled per image, and a pair is represented by the mean of its two image vectors.

\paragraph{Probe and protocol.}
A probe is an $\ell_2$-regularised logistic regression on standardised features reduced to $128$ dimensions by PCA, with the regularisation strength chosen by an inner $5$-fold cross-validation and balanced class weights. All in-domain numbers use $10$-fold cross-validation grouped by HardMatch category, so every test pair comes from a category the probe never saw. We report ROC-AUC with $95\%$ confidence intervals from a cluster bootstrap that resamples whole categories ($1{,}000$ resamples). For transfer, a probe is trained on all HardMatch pairs and applied unchanged to the other datasets.

\paragraph{Baselines from VGGT's own outputs.}
We compare against signals a user of VGGT already has, each with a direction fixed in advance: the lower of the two images' mean point-map confidence, the lower mean depth confidence, image sharpness (variance of the Laplacian), the predicted relative rotation angle, and the ratio of predicted focal lengths. We also train classifiers, with the same cross-validation as the probes, on (i) all confidence statistics and (ii) everything VGGT outputs: confidence statistics, both extrinsics, focal lengths and relative rotation. Baseline (ii) asks whether the probe merely re-reads VGGT's outputs.

\section{Experiments}
\label{sec:experiments}

\subsection{Setup}
\label{sec:setup}

\Cref{tab:datasets} summarises the three datasets. They play different roles. HardMatch is where VGGT fails often enough to study failure in depth, so all probes are trained and analysed there. PhotoTourism and ScanNet-1500 have ground-truth poses but few failures; we use them to ask whether the HardMatch probe transfers, without retraining, to other data and to a label defined directly on the rotation.

\begin{table}[tb]
\centering\small
\caption{\textbf{Datasets.} Ground truth (hand-annotated correspondences, about $11$ per pair, or poses), the failure definition used for each, and VGGT's failure rate on image pairs.}
\label{tab:datasets}
\setlength{\tabcolsep}{3pt}
\begin{tabular}{@{}lrllr@{}}
\toprule
Dataset & Pairs & GT & Failure & Rate \\
\midrule
HardMatch~\cite{nordstrom2026loma} & 903 & corresp. & epi.\ $>50$\,px & $34.9\%$ \\
PhotoTourism~\cite{image-matching-challenge-2022} & 1{,}262 & SfM & $e_R>15^\circ$ & $1.7\%$ \\
ScanNet-1500~\cite{sun2021loftr} & 1{,}500 & RGB-D & $e_{\text{pose}}>15^\circ$ & $7.7\%$ \\
\bottomrule
\end{tabular}
\end{table}

\paragraph{Protocol.}
All probes follow \cref{sec:method}: frozen VGGT, mean-pooled tokens, a linear classifier on 128 PCA components, and cross-validation grouped by HardMatch category, so that every reported number is measured on categories the probe never saw. Unless stated otherwise, the probe reads the last global-attention block (\gl), a choice we fix in advance. Code will be released publicly.

% ---------------------------------------------------------------------------------------------
\subsection{HardMatch: does VGGT encode its own failures?}
\label{sec:hardmatch}

\begin{figure*}[t]
  \centering
  \includegraphics[width=\textwidth]{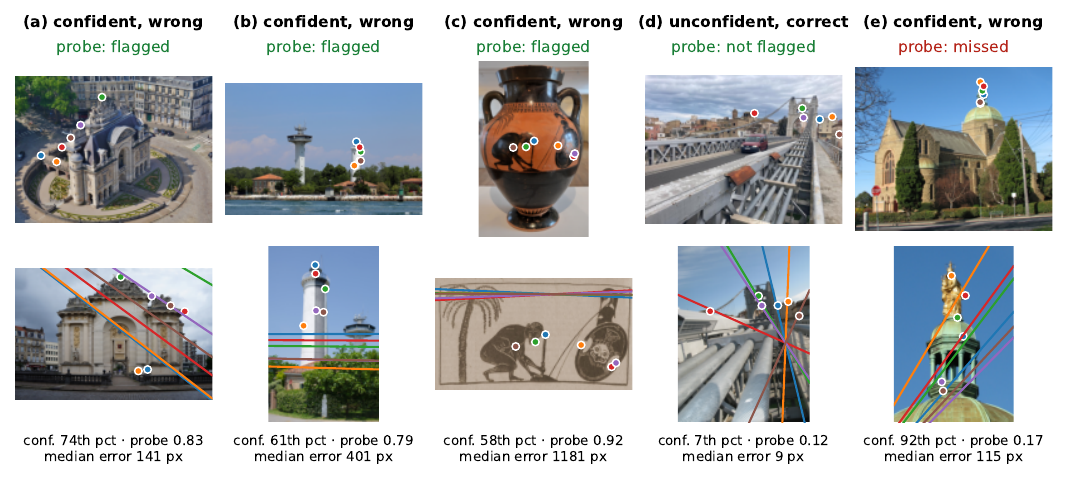}
  \caption{\textbf{The probe catches failures VGGT is confident about.} Top: first image with hand-annotated points. Bottom: second image with the epipolar lines implied by VGGT's own predicted cameras (same colour as the point); for a correct pose each point lies on its line. ``conf.'': percentile of VGGT's point-map confidence among all pairs (higher = more confident); ``probe'': out-of-fold failure probability. (a)--(c) VGGT is confident but wrong (aerial \vs ground view, large viewpoint change, photograph \vs drawing) and the probe flags it. (d) VGGT is unconfident but correct, and the probe does not flag it. (e) A miss: extreme zoom on a confidently wrong pair.}
  \label{fig:qualitative}
\end{figure*}

After the official filtering, HardMatch contains $903$ valid pairs in $97$ categories ($830$ in the official test split), including large viewpoint changes, aerial-to-ground pairs, illumination and season changes, drawings and repeating structure. Judged by the geometry its own cameras imply, VGGT fails on $34.9\%$ of pairs at $50$\,px ($63.7\%$, $51.5\%$ and $25.1\%$ at $10$, $20$ and $100$\,px, respectively), and reaches mAA@10\,px of $19.9\%$ on the official test split. \Cref{fig:qualitative} shows typical pairs.

\paragraph{The features know more than the outputs.}
\Cref{tab:main} compares what a user of VGGT can already observe with what is linearly readable inside it. The most informative output is the point-map confidence ($79.7\%$), followed by depth confidence ($66.5\%$); the predicted pose itself is nearly uninformative (relative rotation $62.9\%$, focal ratio $51.1\%$). A classifier trained on everything VGGT outputs, including both cameras and every confidence statistic, reaches only $77.7\%$, so the outputs contain little failure information beyond the point-map confidence. In contrast, a linear probe on any of the four late layers in \cref{tab:main} reaches $93$--$94\%$, and adding the confidence statistics to the probe leaves it unchanged ($93.5\%$). Thresholded at $0.5$, the probe labels $87.5\%$ of pairs correctly, catching $87.6\%$ of failures at $78.9\%$ precision; VGGT's confidence, thresholded to flag about as many pairs, reaches $74.0\%$ accuracy, catching $68.3\%$ of failures at $61.4\%$ precision.

\begin{table}[tb]
\centering\small
\caption{\textbf{VGGT's features know more about its failures than anything it outputs.} Predicting VGGT's pose failures on HardMatch (failure: median epipolar error $>50$\,px). ROC-AUC (\%) with $95\%$ category-bootstrap CI, and average precision (AP, \%). All learned rows use the same category-grouped cross-validation.}
\label{tab:main}
\setlength{\tabcolsep}{4pt}
\begin{tabular}{@{}lcc@{}}
\toprule
Predictor & ROC-AUC (\%) & AP (\%) \\
\midrule
\multicolumn{3}{@{}l}{\textit{VGGT's own outputs}} \\
\quad Point-map conf.\ (min of pair) & $79.7$ {\scriptsize[76.6, 82.4]} & $60.4$ \\
\quad Depth conf.\ (min of pair) & $66.5$ {\scriptsize[62.8, 69.9]} & $47.5$ \\
\quad Image sharpness (min) & $57.7$ {\scriptsize[53.2, 62.1]} & $44.0$ \\
\quad Predicted rel.\ rotation & $62.9$ {\scriptsize[58.5, 67.1]} & $43.1$ \\
\quad Predicted focal ratio & $51.1$ {\scriptsize[46.6, 55.4]} & $37.1$ \\
\midrule
\multicolumn{3}{@{}l}{\textit{Classifiers trained on VGGT's outputs}} \\
\quad All confidence statistics & $75.2$ {\scriptsize[71.9, 78.7]} & $58.4$ \\
\quad All VGGT outputs & $77.7$ {\scriptsize[73.8, 80.8]} & $59.5$ \\
\midrule
\multicolumn{3}{@{}l}{\textit{Linear probes on frozen features}} \\
\quad ViT block 6 & $59.3$ {\scriptsize[54.6, 64.0]} & $43.8$ \\
\quad ViT block 12 & $64.6$ {\scriptsize[59.9, 69.4]} & $48.7$ \\
\quad ViT block 23 & $65.0$ {\scriptsize[60.3, 69.6]} & $48.3$ \\
\quad Aggregator, last frame-attn & $\mathbf{93.7}$ {\scriptsize[92.0, 95.4]} & $88.4$ \\
\quad Aggregator, last global-attn & $\mathbf{93.5}$ {\scriptsize[91.7, 95.2]} & $87.4$ \\
\quad Camera token & $93.2$ {\scriptsize[91.4, 94.9]} & $87.3$ \\
\quad Camera trunk input & $93.3$ {\scriptsize[91.4, 95.1]} & $\mathbf{88.5}$ \\
\bottomrule
\end{tabular}
\end{table}

The gap has a plausible explanation. VGGT's confidence is trained per pixel to down-weight hard-to-regress points; it describes how certain each depth or point is, not whether the two cameras are placed consistently. The fused features, by contrast, are the very representation from which the camera head regresses the relative pose, and they retain evidence that the two views were not registered consistently, which the camera head discards.

\paragraph{Robustness.}
\Cref{tab:robust} rules out the obvious artefacts. Shuffled labels give chance ($51.6\%$), so the grouped cross-validation does not leak; image size and aspect ratio alone give chance ($49.8\%$), so the padding of the input canvas does not leak the label. Pairs on which VGGT predicts an almost-zero baseline make the fundamental matrix ill-conditioned and could produce spurious failures, but the predicted baseline does not predict failure ($49.9\%$) and removing those pairs leaves the probe at $93.4\%$. The advantage over confidence holds at every threshold, on the official test split, and for every pair representation and PCA size we tried. Combining layers does not help, and combining the three backbone blocks still gives only $64.1\%$.

\begin{table}[tb]
\centering\small
\caption{\textbf{The result is robust to the failure threshold and to every control we tried} (HardMatch, ROC-AUC, \%). Top: failure threshold. Bottom: controls and probe variants (\gl\ unless noted).}
\label{tab:robust}
\setlength{\tabcolsep}{4pt}
\begin{tabular}{@{}rcccc@{}}
\toprule
Threshold & Fail rate & Probe & Point conf. & All outputs \\
\midrule
$10$\,px & $64\%$ & $\mathbf{91.1}$ & $72.8$ & $73.8$ \\
$20$\,px & $51\%$ & $\mathbf{92.9}$ & $76.0$ & $76.6$ \\
$50$\,px & $35\%$ & $\mathbf{93.5}$ & $79.7$ & $77.7$ \\
$100$\,px & $25\%$ & $\mathbf{92.4}$ & $77.8$ & $76.9$ \\
\midrule
\multicolumn{4}{@{}l}{\textit{Controls and probe variants}} & \textit{Probe} \\
\multicolumn{4}{@{}l}{Shuffled labels (control)} & $51.6$ \\
\multicolumn{4}{@{}l}{Image size and aspect only (control)} & $49.8$ \\
\multicolumn{4}{@{}l}{5 folds / official test split only} & $93.4$ / $93.7$ \\
\multicolumn{4}{@{}l}{Without 45 near-zero-baseline pairs} & $93.4$ \\
\multicolumn{4}{@{}l}{All seven layers / frame-attn + global-attn} & $93.2$ / $93.4$ \\
\multicolumn{4}{@{}l}{Concatenated views / mean $+\,|A-B|$} & $93.4$ / $93.7$ \\
\multicolumn{4}{@{}l}{PCA $32$ / $64$ / $256$ dims} & $93.4$ / $93.7$ / $93.5$ \\
\multicolumn{4}{@{}l}{ViT blocks 6+12+23 combined} & $64.1$ \\
\bottomrule
\end{tabular}
\end{table}

\begin{figure*}[t]
  \centering
  \includegraphics[width=\textwidth]{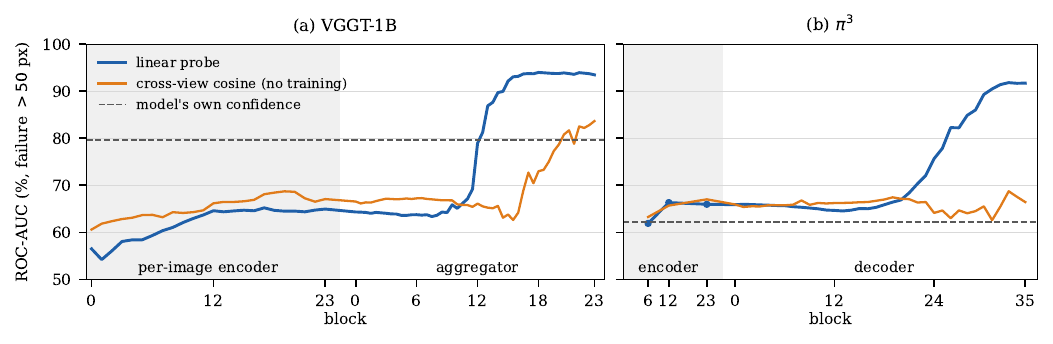}
  \caption{\textbf{Failure awareness emerges abruptly in the middle of the network, in both models.} ROC-AUC for predicting VGGT's pose failures on HardMatch from every block of (a) VGGT and (b) $\pi^3$, with a linear probe (blue) and with one training-free number, the cosine similarity between the two views' pooled features (orange). Dashed: the model's own confidence. Shaded: the per-image encoder. The probe is flat through the encoder and the first half of the cross-view layers, then rises within a few blocks; only in VGGT does the training-free similarity follow.}
  \label{fig:layers}
\end{figure*}

\paragraph{Where the signal emerges.}
To locate the signal we probe every block of the network (\cref{fig:layers}a). The per-image encoder never exceeds $65.2\%$: what makes a pair hard is not visible in either image alone. More surprisingly, the first half of the aggregator adds nothing either, although it already alternates frame and global attention: through block $10$ the probe stays between $63\%$ and $66\%$. It then rises abruptly, from $69\%$ after block $11$ to $81\%$ after block $12$, $88\%$ after block $13$ and $93\%$ after block $15$, and stays at $93$--$94\%$ to the last block (peak $94.0\%$, block $18$). The failure signal is therefore not a by-product of the two views meeting; it is computed in a narrow band of middle aggregator blocks. A training-free reading appears later still. The cosine similarity between the two views' pooled features stays at the encoder's level ($61$--$69\%$) through block $15$ and then climbs to $83.7\%$ at the last block, above VGGT's own confidence: in its final layers, VGGT's representations of the two views stop agreeing when its pose is wrong. At the camera token this similarity falls back to $71.0\%$ while a trained probe stays at $93.2\%$; the information is still present, but no longer as a simple disagreement.

\begin{table}[tb]
\centering\small
\caption{\textbf{A second model, $\pi^3$, shows the same gap between its features and its confidence.} HardMatch, same pairs, failure definition and protocol. ROC-AUC (\%) unless stated; ``most confident half'': the half of the pairs on which the model's own confidence is highest.}
\label{tab:pi3}
\setlength{\tabcolsep}{4pt}
\begin{tabular}{@{}lcc@{}}
\toprule
 & VGGT & $\pi^3$ \\
\midrule
Failure rate ($>50$\,px) & $34.9\%$ & $34.6\%$ \\
mAA@10\,px (test split) & $19.9$ & $18.4$ \\
\midrule
Own confidence (min of pair) & $79.7$ & $62.2$ \\
All outputs (learned) & $77.7$ & $76.9$ \\
Probe, best of encoder blocks 6/12/23 & $65.0$ & $66.3$ \\
Probe, last attention block & $\mathbf{93.5}$ & $\mathbf{91.7}$ \\
Probe, camera head & $93.2$ & $88.7$ \\
Cross-view cosine, best block & $83.7$ & $68.7$ \\
\midrule
Most confident half: failure rate & $12.2\%$ & $27.3\%$ \\
\quad own confidence & $69.9$ & $52.8$ \\
\quad probe & $\mathbf{90.0}$ & $\mathbf{92.7}$ \\
\bottomrule
\end{tabular}
\end{table}

\paragraph{A second model: $\pi^3$.}
$\pi^3$~\cite{wang2026pi} also alternates frame-wise and global attention, but has no reference view and predicts camera-to-world poses and local point maps, from which we recover the focal length by least squares. We run it on the same pairs with the same failure definition. It fails on $34.6\%$ of them (mAA@10 $18.4\%$ on the test split) and agrees with VGGT's failure label on $81\%$. \Cref{tab:pi3} repeats the main comparison. $\pi^3$'s own confidence is a weaker failure signal than VGGT's ($62.2\%$), and on its most confident half, where $27\%$ of pairs still fail, it is at chance ($52.8\%$). A probe on its last decoder block nonetheless reaches $91.7\%$, against $76.9\%$ for a classifier trained on all of its outputs, and $92.7\%$ on that most confident half. The layer profile repeats as well (\cref{fig:layers}b): flat through the encoder and the first $21$ decoder blocks ($\leq 67\%$), then a steep rise to $92\%$ by block $33$. One difference is instructive: in $\pi^3$ the training-free cosine never rises ($\leq 69\%$). Failure is linearly decodable in both models, but the simple reading that the two views stop agreeing is specific to VGGT.

\begin{figure}[tb]
  \centering
  \includegraphics[width=\linewidth]{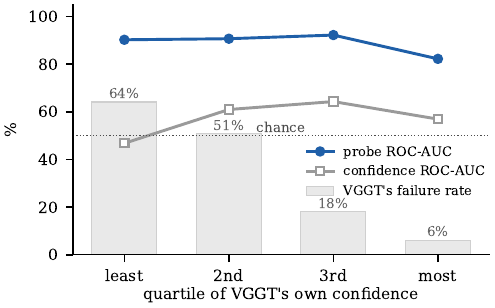}
  \caption{\textbf{Within each level of confidence, VGGT's confidence is near chance; the probe is not.} HardMatch pairs split into quartiles of VGGT's point-map confidence. Bars: VGGT's failure rate. Lines: ROC-AUC for predicting failure within each quartile.}
  \label{fig:quartiles}
\end{figure}

\paragraph{Confidently wrong predictions.}
Confidence and failure are correlated at the level of the whole benchmark, but how much does confidence tell us once we already know it? We split the pairs into quartiles of point-map confidence. The failure rate falls from $64\%$ in the least confident quartile to $6\%$ in the most confident, yet \emph{within} each quartile confidence ranks failures close to chance (\cref{fig:quartiles}) (AUC $47\%$, $61\%$, $64\%$, $57\%$) while the probe stays high ($90\%$, $91\%$, $92\%$, $82\%$). On the most confident half of the pairs, where $12.2\%$ still fail, the probe reaches $90.0\%$ [$85.4$, $94.2$] against $69.9\%$ [$61.2$, $77.8$]. These are the failures a downstream system cannot see. \Cref{fig:qualitative} shows examples: an aerial-to-ground pair, a large viewpoint change and a photograph matched to a drawing, all of which VGGT places confidently but wrongly and the probe flags, together with a correct pair on which confidence raises a false alarm and an extreme-zoom pair the probe misses.

\begin{figure}[tb]
  \centering
  \includegraphics[width=\linewidth]{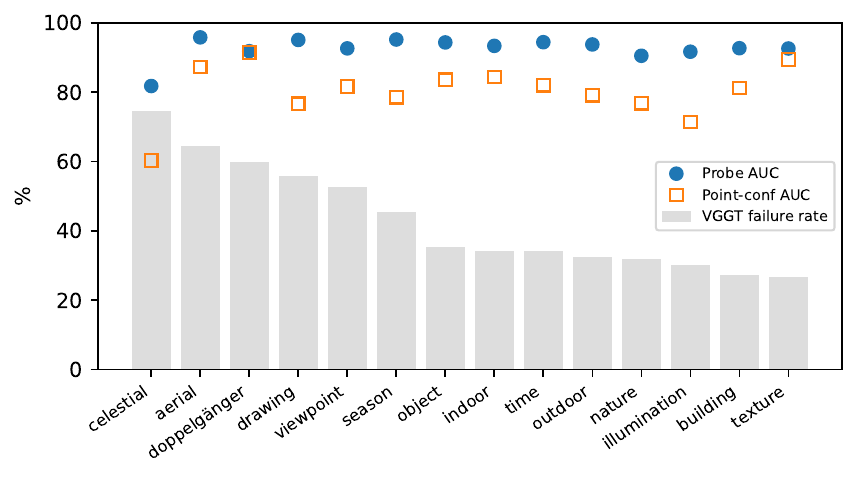}
  \caption{\textbf{The probe beats VGGT's confidence in every HardMatch category but one.} Bars: VGGT's failure rate. Markers: AUC of the probe (filled) and of point-map confidence (open).}
  \label{fig:categories}
\end{figure}

\paragraph{Per category.}
Failure rates vary widely across HardMatch's categories, from $27\%$ on building and texture pairs to $74\%$ on celestial ones (\cref{fig:categories}). The probe exceeds VGGT's confidence in every category with enough pairs except doppelg{\"a}ngers, where both reach $92\%$; the largest gains are on the categories where appearance changes most, namely celestial ($82\%$ \vs $60\%$), drawing ($95\%$ \vs $77\%$) and illumination ($92\%$ \vs $71\%$).

\begin{figure}[tb]
  \centering
  \includegraphics[width=\linewidth]{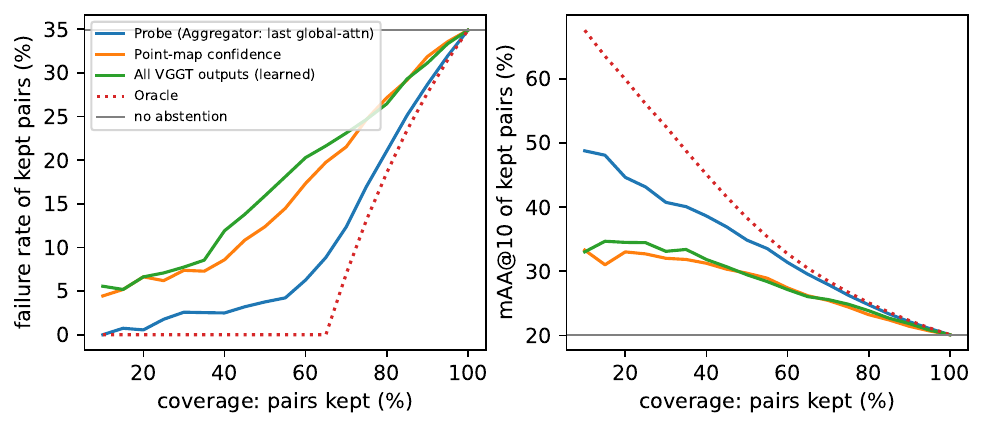}
  \caption{\textbf{Abstaining on the pairs the probe distrusts removes most of VGGT's failures.} Keeping the pairs ranked most reliable by each score: failure rate (left) and mAA@10\,px (right) of the kept pairs. Oracle: ranked by true error.}
  \label{fig:selective}
\end{figure}

\paragraph{Using the signal: selective prediction.}
A system that can abstain on some pairs, for example by requesting another view or falling back to a slower pipeline, needs a ranking rather than a calibrated probability. Keeping the $50\%$ of pairs the probe trusts most yields a $3.8\%$ failure rate and mAA@10 of $34.9\%$; ranking by VGGT's confidence yields $12.4\%$ and $29.7\%$, and a perfect oracle $0\%$ and $38.3\%$ (\cref{fig:selective}). Over all coverages, the area under the risk--coverage curve is $0.096$ for the probe, $0.154$ for confidence, $0.169$ for all outputs and $0.070$ for the oracle, so the probe closes about $70\%$ of the gap between VGGT's confidence and a perfect ranking.

\paragraph{A negative result: routing to a matcher.}
We also tried to route each pair either to VGGT or to the dedicated matcher LoMa~\cite{nordstrom2026loma}. On HardMatch, LoMa (mAA@10 $50.8\%$ on the test split) almost never does worse than VGGT ($19.9\%$): even an oracle that always picks the better of the two reaches only $52.1\%$, and a learned router that uses the probe reaches $51.1\%$. With so little complementarity there is nothing for the probe to exploit, and we make no routing claim.

\FloatBarrier
% ---------------------------------------------------------------------------------------------
\subsection{PhotoTourism: transfer to rotation failures}
\label{sec:pt}

HardMatch defines failure through epipolar error, which mixes rotation, translation and intrinsics errors, and its probe is trained on HardMatch's own categories. PhotoTourism~\cite{image-matching-challenge-2022} lets us test both limitations at once: its poses come from large-scale structure-from-motion, so failure can be defined on the rotation alone, and none of its images were used to train the probe. We sample $1{,}262$ pairs from $16$ landmarks, stratified into four bins of ground-truth co-visibility, and apply the HardMatch probe unchanged.

\paragraph{VGGT rarely fails on landmark photographs.}
Rotation errors above $15^\circ$ occur on only $1.7\%$ of the pairs, and almost all of them on pairs that barely overlap: $6.6\%$ of pairs with co-visibility below $0.02$ fail, against at most $0.3\%$ in every other bin (\cref{tab:pt}, top). At the looser threshold of $5^\circ$, $6.8\%$ of pairs fail. These pairs are much easier than HardMatch, which is why we use PhotoTourism as a transfer test rather than as a training set.

\begin{table}[tb]
\centering\small
\caption{\textbf{The HardMatch probe transfers to rotation failures on PhotoTourism, beyond co-visibility.} Top: VGGT's rotation failures by co-visibility. Bottom: ROC-AUC (\%) for rotation failures of the HardMatch probe (no retraining), VGGT's confidence, ground-truth co-visibility (not available at test time) and a probe trained on PhotoTourism itself (scene-grouped CV). ``low covis.'': pairs with co-visibility $<0.02$ only.}
\label{tab:pt}
\setlength{\tabcolsep}{2.5pt}
\begin{tabular}{@{}lcccc@{}}
\toprule
Co-visibility & $<0.02$ & $0.02$--$0.1$ & $0.1$--$0.3$ & $>0.3$ \\
\midrule
Pairs & 302 & 320 & 320 & 320 \\
$e_R>15^\circ$ & $6.6\%$ & $0.3\%$ & $0.0\%$ & $0.3\%$ \\
\midrule
 & \multicolumn{2}{c}{all pairs} & \multicolumn{2}{c}{low covis.} \\
\cmidrule(lr){2-3}\cmidrule(l){4-5}
 & $e_R>10^\circ$ & $e_R>15^\circ$ & $e_R>10^\circ$ & $e_R>15^\circ$ \\
\midrule
\#failures & 25 & 22 & 21 & 20 \\
\textbf{HardMatch probe} & $\mathbf{94.6}$ & $\mathbf{97.5}$ & $\mathbf{86.9}$ & $\mathbf{89.5}$ \\
VGGT confidence & $88.5$ & $90.6$ & $79.5$ & $79.2$ \\
GT co-visibility & $87.0$ & $89.8$ & $74.3$ & $75.7$ \\
In-domain probe & $95.1$ & $89.1$ & -- & -- \\
\bottomrule
\end{tabular}
\end{table}

\paragraph{The HardMatch probe transfers.}
Without seeing a single PhotoTourism image, the probe predicts rotation failures with AUC $94.6\%$ at $10^\circ$ and $97.5\%$ at $15^\circ$ ($87.1\%$ at $5^\circ$), above VGGT's confidence at every threshold and on par with a probe trained on PhotoTourism itself (\cref{tab:pt}, bottom). The score is also related to the size of the error, not just to a threshold: over all $1{,}262$ pairs its Spearman correlation with the log rotation error is $0.43$, against $0.33$ for confidence. The intervals are wide because there are few failures ($25$ at $10^\circ$), but the ordering is the same at every threshold.

\paragraph{It is not just co-visibility.}
Because the failures concentrate on barely overlapping pairs, a probe that merely detected low overlap, as Co-VGGT's~\cite{ziliotto2026covggt} does by design, would also score well. Three observations argue against this explanation. First, ground-truth co-visibility itself, which a deployed system would not have, predicts the failures less well than the probe ($87.0\%$ and $89.8\%$). Second, \emph{among only the barely overlapping pairs}, where co-visibility is nearly constant, the probe still separates failures from successes ($86.9\%$ and $89.5\%$) better than confidence ($79\%$) and co-visibility ($74$--$76\%$). Third, with co-visibility partialled out, the probe's rank correlation with the rotation error remains $0.23$, twice that of confidence ($0.11$). The probe detects whether VGGT got the rotation wrong, not merely whether the views overlap.

\paragraph{Multi-view graphs: no downstream gain.}
We also built ten $16$-image graphs from PhotoTourism scenes and ran VGGT on all $120$ pairs of each. On these $1{,}200$ edges the HardMatch probe again separates rotation failures ($e_R>15^\circ$, $1.75\%$ of edges) better than confidence ($98.6\%$ \vs $92.7\%$). We then asked whether discarding distrusted edges improves rotation averaging. Robust averaging over all edges already reduces the mean absolute-rotation error from $1.27^\circ$ (least squares) to $0.70^\circ$; dropping the $10\%$ of edges the probe distrusts most gives $0.66^\circ$, dropping by confidence $0.67^\circ$, and even an oracle that removes every edge whose true error exceeds $15^\circ$ gives $0.67^\circ$. With fewer than $2\%$ of edges failing, robust averaging already absorbs them; showing a downstream benefit needs graphs with far denser failures than these.

\begin{table}[tb]
\centering\small
\caption{\textbf{More views remove most failures; the few left are hard to rank for both the probe and confidence} (PhotoTourism, $96$ sets of $16$ images). Top: failure rate of the same $2{,}304$ pairs when VGGT sees $2$, $4$, $8$ or $16$ views per forward pass. Bottom: all pairs of each setting; ROC-AUC (\%) for rotation errors above $5^\circ$ of the HardMatch probe (trained on two-view pairs, no retraining) and of VGGT's confidence. With $28$--$83$ failures per setting, the $95\%$ confidence intervals span more than $40$ points.}
\label{tab:multiview}
\setlength{\tabcolsep}{4pt}
\begin{tabular}{@{}lcccc@{}}
\toprule
Views per forward pass & $2$ & $4$ & $8$ & $16$ \\
\midrule
Same pairs: $e_R>5^\circ$ & $3.6\%$ & $1.4\%$ & $0.5\%$ & $0.3\%$ \\
Same pairs: $e_R>15^\circ$ & $0.9\%$ & $0.3\%$ & $0.0\%$ & $0.0\%$ \\
\midrule
All pairs & 2{,}304 & 2{,}304 & 5{,}376 & 11{,}520 \\
\quad failures ($e_R>5^\circ$) & 83 & 33 & 28 & 35 \\
\quad HardMatch probe (2-view) & $81.5$ & $82.6$ & $74.2$ & $86.3$ \\
\quad VGGT confidence & $91.2$ & $90.4$ & $92.9$ & $82.1$ \\
\bottomrule
\end{tabular}
\end{table}

\paragraph{More than two views.}
HardMatch defines its ground truth per pair, so our main study uses pairs. To test whether this matters, we sample $96$ sets of $16$ images from the PhotoTourism scenes and run VGGT on each set as one $16$-view pass, as two $8$-view passes, as four $4$-view passes, and on every pair inside the $4$-view groups, so that the same $2{,}304$ pairs are predicted with $2$, $4$, $8$ and $16$ views (\cref{tab:multiview}). Extra views fix most failures: the share of these pairs with a rotation error above $5^\circ$ falls from $3.6\%$ with two views to $1.4\%$, $0.5\%$ and $0.3\%$ with $4$, $8$ and $16$, and with $8$ or more views none exceeds $10^\circ$. The probe trained on two-view HardMatch pairs still ranks the remaining failures well above chance without retraining ($74$--$86\%$ for $4$--$16$ views), and VGGT's own confidence does comparably ($82$--$93\%$); with $28$--$35$ failures per setting, neither is clearly better. Even with two views, confidence is ahead for small errors (above $5^\circ$: $91.2\%$ \vs $81.5\%$) and the probe for larger ones (above $10^\circ$: $95.7\%$ \vs $93.6\%$). Extra views thus mainly remove the failures the probe is meant to catch; failure awareness for many-view inputs needs a benchmark on which such inputs still fail.

\FloatBarrier
% ---------------------------------------------------------------------------------------------
\subsection{ScanNet-1500: failure modes differ}
\label{sec:scannet}

ScanNet-1500~\cite{dai2017scannet,sun2021loftr} is the standard indoor benchmark for relative pose: $1{,}500$ pairs from $100$ rooms, with poses from RGB-D reconstruction. Indoor scenes add low texture, repeated structure and, unlike PhotoTourism, many pairs taken from nearby positions.

\paragraph{VGGT's indoor failures are translation failures.}
VGGT's rotations on ScanNet are accurate (median error $1.2^\circ$; only $2.1\%$ of pairs exceed $5^\circ$). Its pose errors come almost entirely from the direction of the translation: $113$ of the $115$ pairs with pose error above $15^\circ$ have a rotation error below $5^\circ$. These failures are concentrated on short baselines (\cref{tab:scannet}, top): the translation direction is wrong by more than $15^\circ$ on $47\%$ of pairs whose cameras are less than $10$\,cm apart, but on fewer than $1\%$ of pairs more than $40$\,cm apart. This is a known property of two-view geometry rather than a peculiarity of VGGT: when the cameras barely move, the direction of their small displacement is weakly constrained by the images.

\begin{table}[tb]
\centering\small
\caption{\textbf{On ScanNet, the probe transfers to rotation failures but not to short-baseline translation failures.} Top: VGGT's errors by true camera baseline; translation-direction errors dominate at short baselines. Bottom: ROC-AUC (\%) of the HardMatch probe (no retraining), VGGT's confidence and a probe trained on ScanNet itself (scene-grouped CV), by failure type. $e_{\text{pose}}=\max(e_R,e_t)$.}
\label{tab:scannet}
\setlength{\tabcolsep}{4pt}
\begin{tabular}{@{}lcccc@{}}
\toprule
True baseline & $<10$\,cm & $10$--$20$\,cm & $20$--$40$\,cm & $>40$\,cm \\
\midrule
Pairs & 83 & 173 & 292 & 952 \\
$e_t>15^\circ$ & $47.0\%$ & $27.2\%$ & $6.8\%$ & $0.9\%$ \\
$e_R>5^\circ$ & $0.0\%$ & $0.0\%$ & $0.3\%$ & $3.3\%$ \\
\midrule
Failure & \#fail & HM probe & Confidence & In-domain \\
\midrule
$e_R>5^\circ$ & 32 & $79.5$ & $58.7$ & $\mathbf{82.3}$ \\
$e_{\text{pose}}>10^\circ$ & 202 & $47.6$ & $61.0$ & $\mathbf{90.5}$ \\
$e_{\text{pose}}>15^\circ$ & 115 & $44.2$ & $63.6$ & $\mathbf{91.0}$ \\
\bottomrule
\end{tabular}
\end{table}

\paragraph{Transfer depends on the failure mode.}
The few rotation failures behave differently: they occur at the longest baselines ($3.3\%$ of pairs above $40$\,cm, none below $20$\,cm), the regime closest to HardMatch's wide-baseline pairs. For these, the HardMatch probe transfers: $79.5\%$ against $58.7\%$ for confidence, close to a probe trained on ScanNet itself ($82.3\%$). For the translation-dominated failures it does not ($44$--$48\%$), and VGGT's confidence is only marginally better ($61$--$64\%$). Yet a probe trained on ScanNet detects these failures with AUC $90.5$--$91.0\%$. VGGT's fused features therefore encode the short-baseline failures as well, but along a direction that a probe trained on HardMatch's gross failures does not learn. Failure awareness is not a single scalar inside the model: one linear readout carries across datasets for wrong orientation, and ill-conditioned translation calls for its own.

\FloatBarrier

\section{Limitations}
\label{sec:limitations}
\emph{Two models.} We study VGGT and $\pi^3$; larger or newer models (VGGT-$\Omega$, MapAnything) may behave differently, and the training-free cosine reading already differs between the two.
\emph{Labels.} HardMatch has no ground-truth poses, so failure is defined through the epipolar error of about $11$ annotated correspondences per pair, which mixes rotation, translation and intrinsics errors; the ground-truth datasets confirm the result for rotation but have few failures (PhotoTourism: $22$ above $15^\circ$).
\emph{Two views.} We study image pairs, because HardMatch defines its ground truth per pair. With more views, failures on PhotoTourism become rare ($0.3\%$ with $16$ views) and the two-view probe is no longer clearly better than VGGT's confidence on them; filtering edges by the probe also did not improve rotation averaging (\cref{sec:pt}).
\emph{Supervision.} The probe is trained with failure labels, whereas VGGT's confidence is free. We therefore compare against classifiers trained on VGGT's outputs under the same protocol, and test the probe on data it was not trained on.
\emph{Failure modes.} A single probe does not cover short-baseline translation errors (\cref{sec:scannet}).

\section{Conclusion}
\label{sec:conclusion}
VGGT knows more about its own failures than it reports.
A linear readout of its frozen features predicts when its relative pose is wrong far better than its own confidence or anything else it outputs, including when it is confident; it supports selective prediction and transfers to rotation failures on other datasets.
The signal emerges abruptly in the middle of the network, in both VGGT and $\pi^3$.
Costing one linear layer on features the model already computes, it is a practical reliability signal for systems built on feed-forward reconstruction, and evidence that self-knowledge, studied extensively in language models, is also present in 3D foundation models.

{
    \small
    \bibliographystyle{ieeenat_fullname}
    \bibliography{main}
}
\end{document}